\documentclass{article}
\usepackage{ijcai26}

\usepackage{times}
\usepackage{soul}
\usepackage{url}
\usepackage[hidelinks]{hyperref}
\usepackage[utf8]{inputenc}
\usepackage[small]{caption}
\usepackage{graphicx}
\usepackage{amsmath}
\usepackage{amsthm}
\usepackage{bm}
\usepackage{booktabs}
\usepackage{algorithm}
\usepackage{algorithmic}
\usepackage[switch]{lineno}
\usepackage{amssymb}    
\usepackage{booktabs, tabularx, array}
\newcolumntype{L}[1]{>{\raggedright\arraybackslash}p{#1}}

\title{Wasserstein Evolution : Evolutionary Optimization as Phase Transition}

\author{
Kaichen Ouyang$^1$
\and
Mingyang Yu$^2$
\and
Zong Ke$^3$
\and
Junbo Jacob Lian$^4$
\and
Shengwei Fu$^5$
\and
Xiaoyang Hao$^6$
\and
Shengju Yu$^7$
\and
Dayu Hu$^6$\\
\affiliations
$^1$University of Science and Technology of China\\
$^2$Nankai University\\
$^3$National University of Singapore\\
$^4$Northwestern University\\
$^5$Guizhou University\\
$^6$Northeastern University\\
$^7$Hong Kong Baptist University\\
\emails
oykc@mail.ustc.edu.cn,
hudy@bmie.neu.edu.cn
}
\usepackage{fancyhdr}
\begin{document}
\maketitle

\thispagestyle{fancy}
\fancyhf{} 
\renewcommand{\headrulewidth}{0pt} 
\renewcommand{\footrulewidth}{0.4pt} 
\renewcommand{\footrule}{\hrule width 4.5cm\relax} 
\fancyfoot[L]{\small Preprint. Under review.} 

\pagestyle{empty}

\begin{abstract}
Evolutionary algorithms (EAs) serve as powerful black-box optimizers inspired by biological evolution. However, most existing EAs predominantly focus on heuristic operators such as crossover and mutation, while usually overlooking underlying physical interpretability such as statistical mechanics and thermosdynamics. This theoretical void limits the principled understanding of algorithmic dynamics, hindering the systematic design of evolutionary search beyond ad-hoc heuristics. To bridge this gap, we first point out that evolutionary optimization can be conceptually reframed as a physical phase transition process. Building on this perspective, we establish the theoretical grounds by modeling the optimization dynamics as a Wasserstein gradient flow of free energy. Consequently, a robust and interpretable solver named Wasserstein Evolution (WE) is proposed.  WE mathematically frames the trade-off between exploration and exploitation as a competition between potential gradient forces and entropic forces. This formulation guarantees  convergence to the Boltzmann distribution, thereby minimizing free energy and maximizing entropy, which promotes highly diverse solutions. Extensive experiments on complex multimodal and physical potential functions demonstrate that WE achieves superior diversity and stability compared to established baselines.

\end{abstract}

\section{Introduction}
\begin{figure}[t]
	\centering
	\includegraphics[width=\linewidth]{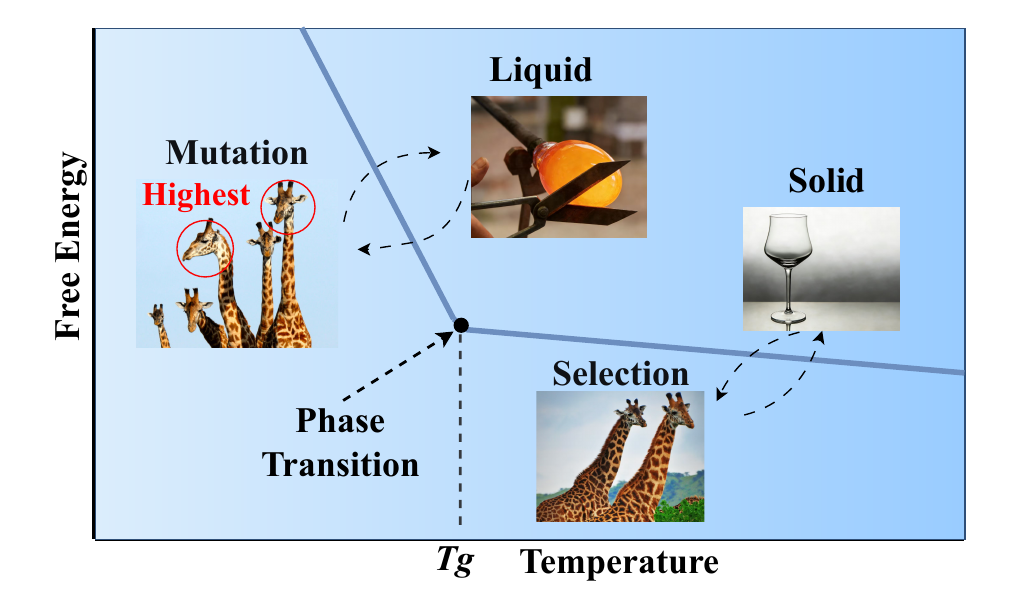}
	\caption{A unified view of phase transitions and evolution. The transition from a liquid to a solid mirrors the shift from biological exploration to exploitation. The glass transition temperature ($T_g$) marks the threshold at which the system switches from mutation-like, entropic exploration to selection-like, potential-driven exploitation.}
	\label{fig:PT}
\end{figure}

Phase transition behavior is commonly observed in complex systems, including glass transitions and ferromagnetic materials\cite{hohenberg1977theory,schoenholz2016structural,schultz1964two}.Such behavior reflects an inherent competition between energetic driving forces, represented by potential gradients, and entropic forces\cite{pachter2024entropy}. Potential gradients drive the system toward lower energy configurations, thereby promoting order. By contrast, entropic forces favor access to a larger set of microscopic configurations in phase space, thereby increasing disorder. The balance between these competing influences guides the system toward a stable free energy minimum. This parallel between order–disorder dynamics and the trade-off between exploration and exploitation in evolutionary algorithms (EAs) motivates the reformulation of these algorithms within a physical framework \cite{vcrepinvsek2013exploration}. Figure~\ref{fig:PT} illustrates the correspondence between phase transition processes and evolutionary dynamics.

The evolutionary process can be understood as an adaptive response of organisms to their environment. Wright’s concept of the fitness landscape captures biological adaptation through an analogy with potential energy in physics\cite{wright1932roles}. In this view, biodiversity can be quantified by entropy, which reflects the number of accessible microscopic configurations in genetic space. Accordingly, evolution may be viewed as a shift from mutation-driven exploration of genotypes to selection-driven retention of advantageous genes. This shift can be interpreted as a phase transition from an entropy-dominated regime to a potential-dominated regime, reflecting the fundamental duality of evolution as a transition from disorder to order\cite{ao2005laws,marinari1992simulated}.

The balance between exploration and exploitation is a central challenge in existing EAs \cite{tan2009balancing,hong2024protein,si2023linear}. Conventional approaches face two key limitations: (1) exploration and exploitation are often intertwined through shared algorithmic parameters that are tuned using empirical feedback, leading to ad hoc manual adjustment and making principled control difficult; and (2) Many methods target a single optimum, which often drives the system to converge to a single point, causing representation collapse and reducing solution diversity\cite{ojha2022assessing,tanabe2019reviewing,karafotias2014parameter}. 

To address these issues, we reinterpret evolutionary optimization as a phase transition in statistical physics and propose Wasserstein Evolution (WE), an algorithm derived from the Wasserstein gradient flow of free energy. In WE, minimizing free energy drives the system toward an equilibrium between potential energy and entropy, thereby adaptively balancing exploration and exploitation. Moreover, WE replaces the conventional single-point convergence objective with convergence to a Boltzmann distribution, which naturally supports the discovery of diverse solutions. Through extensive experiments comparing WE with multiple well-established EAs on ten benchmark functions, which include five classical multimodal functions, five real-world physical potential functions, and Schwefel 2.22 with its 11 transformed variants, WE consistently achieves the highest average entropy, indicating superior solution diversity. Further numerical visualizations demonstrate its characteristic convergence to the Boltzmann distribution. By following the Wasserstein gradient flow, WE minimizes free energy to reach a physically stable equilibrium. Our main contributions are summarized as follows:

\begin{itemize}
\item \textit{Framework:} We investigate the intrinsic duality between phase transitions and EAs. By explicitly modeling evolutionary dynamics within a statistical-physics framework, this work provides a rigorous, physically principled explanation of the exploration–exploitation trade-off.

\item \textit{Algorithm:} We propose a novel Wasserstein evolution algorithm derived from the Wasserstein gradient flow of free energy. Unlike existing EAs, WE adaptively balances exploration and exploitation through free-energy minimization. Moreover, it converges to a Boltzmann distribution rather than a single optimal solution, thereby promoting solution diversity.

\item \textit{Evaluation:} Extensive experiments on ten benchmark functions, including physical potential functions and multimodal functions, demonstrate that WE achieves superior solution diversity with the highest average entropy and effectively mitigates representation collapse.
\end{itemize}

\section{Background}
The proposed WE framework reinterprets the update process of EAs from the perspective of phase transitions. To establish a clear research motivation for the WE framework, we review prior studies on EAs and phase transitions.

\noindent \textbf{Evolutionary Algorithms.} EAs constitute a family of population-based metaheuristic optimization methods inspired by biological evolution. The earliest and most classical form of evolutionary algorithm is the Genetic Algorithm (GA), introduced by Holland in 1992\cite{holland1992genetic}. GA mimics natural selection by employing operators such as selection, crossover, and mutation to evolve a population of candidate solutions toward optimal solutions. Subsequently, the EA framework has expanded to include several major variants, most notably Differential Evolution (DE)\cite{storn1997differential} and Evolution Strategies (ES)\cite{beyer2002evolution}. DE operates through vector-based mutation and crossover operations. Representative algorithms include the Self-Adaptive Differential Evolution algorithm (SaDE)\cite{qin2005self} and Adaptive Differential Evolution with Optional External Archive (JADE)\cite{zhang2009jade}. In contrast, ES emphasizes self-adaptive mutation and deterministic selection mechanisms, often modeling mutation step sizes using covariance matrices. Representative examples include the well-known Covariance Matrix Adaptation Evolution Strategy (CMA-ES)\cite{hansen2001completely,hansen2003reducing} and Natural Evolution Strategy (NES)\cite{wierstra2014natural,glasmachers2010exponential}.

Despite these developments, the core of evolutionary algorithm design remains focused on managing the balance and transition between exploration and exploitation. In the early stages, algorithms emphasize exploration to broadly sample the search space and mitigate the risk of premature convergence. As evolution progresses, the emphasis gradually shifts toward exploitation, enabling more refined searches within promising regions and guiding the algorithm toward high-quality solutions. Effectively controlling this adaptive trade-off between global exploration and local refinement remains a fundamental challenge in evolutionary computation.

\noindent \textbf{Phase Transition.} Phase transitions in statistical physics describe transformations between distinct states of matter. Classical lattice models, such as the Ising model and spin glass models, have been widely used to investigate these phenomena\cite{cipra1987introduction,binder1986spin}. To capture more complex behaviors, researchers have further developed models such as glass transition and jamming models\cite{berthier2011theoretical,liu1998jamming,o2003jamming,o2002random}.

In recent decades, the study of phase transitions has extended beyond physical systems, with related theories increasingly applied to the analysis of complex systems in artificial intelligence.Classic examples include Hopfield networks and Boltzmann machines\cite{ackley1985learning,hopfield1982neural}, which are directly inspired by statistical physics models exhibiting phase transition behavior, particularly those related to spin glasses. More recently, this connection has deepened through the development of statistical-physics-based theories aimed at explaining phenomena in deep learning. Representative studies include linking neural scaling laws to phase transition theory, interpreting grokking as a delayed generalization transition within statistical physics frameworks, and analyzing representation smoothing in graph neural networks as a form of dynamical phase transition\cite{kaplan2020scaling,liu2022towards,oono2019graph}.

Phase transition processes are often associated with the minimization of free energy, which is governed by two principal components: potential gradient forces and entropic forces. In a similar spirit, Darwinian dynamics models evolution as a process driven by selection and variation forces, corresponding respectively to exploitation and exploration behaviors in EAs\cite{ao2008emerging}. This conceptual alignment suggests the feasibility of reinterpreting evolutionary algorithms from the perspective of phase transitions.

\noindent \textbf{Bridging the Gap.}  Previous research has rarely examined EAs from a phase transition perspective, resulting in limited underlying physical interpretability. Meanwhile, existing EAs largely rely on metaphor-driven frameworks and function-evaluation-based designs, often requiring ad-hoc manual parameter tuning. In contrast, our work reframes the dynamics of EAs as a phase transition process, thereby adaptively balancing exploration and exploitation. Specifically, this balance is achieved by performing evolutionary optimization via Wasserstein free energy gradient flows, governed by the interplay between potential gradient forces and entropic forces.

\begin{figure}[t]
	\centering
	\includegraphics[width=\linewidth]{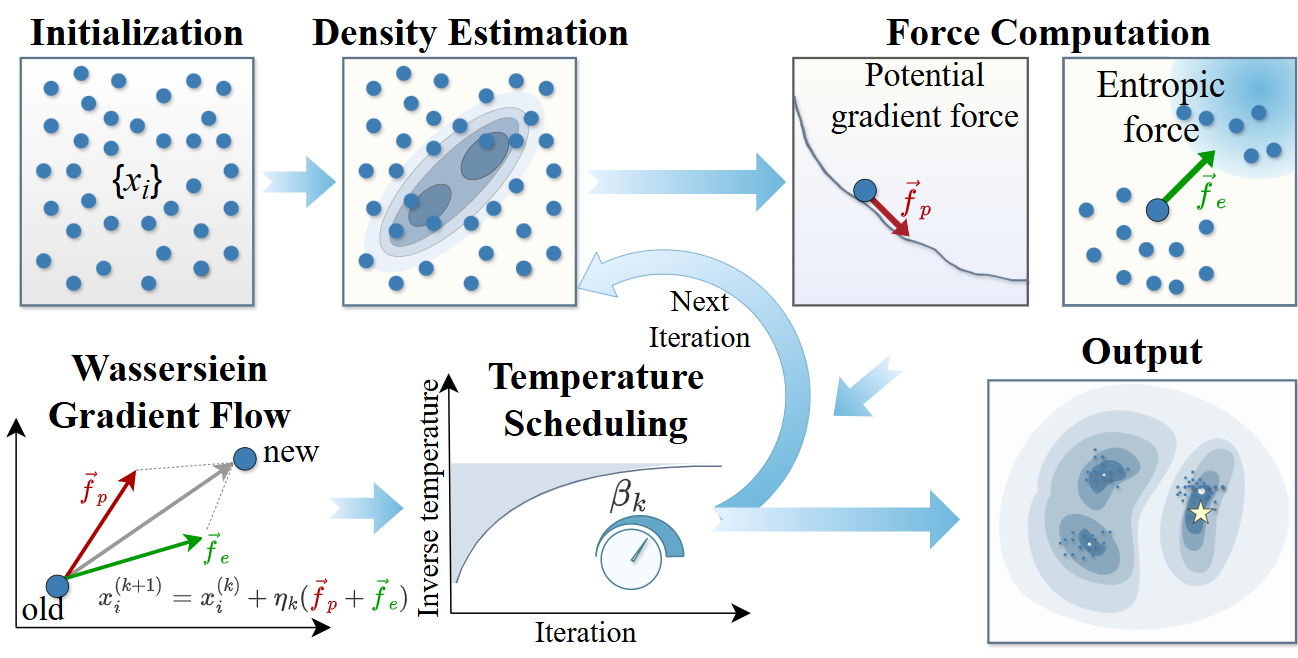}
	\caption{The framework of Wasserstein Evolution. It iteratively minimizes free energy by balancing the potential gradient force (exploitation) and the entropic force (exploration). At each step, the population density is modeled via Kernel Density Estimation (KDE) to compute the resultant force, which drives individuals toward the equilibrium Boltzmann distribution.}
	\label{fig:fra}
\end{figure}

\section{Evolutionary Optimization as Phase Transition}
\subsection{Preliminary}
The theoretical basis of Wasserstein Evolution is grounded in optimal transport theory and gradient flows in metric spaces\cite{otto2001geometry,jordan1998variational,benamou2000computational,chizat2018global}. In this section, we introduce the relevant concepts.

\noindent \textbf{Wasserstein Space and Gradient Flows.}  
We consider the space $\mathcal{P}_2(\mathbb{R}^d)$ of probability measures on $\mathbb{R}^d$ with finite second moments. The $W_2$ Wasserstein distance $W_2(\mu,\nu)$ between $\mu,\nu\in\mathcal{P}_2(\mathbb{R}^d)$ is defined as the minimal quadratic cost of transporting mass from $\mu$ to $\nu$. For a functional $\mathcal{F}:\mathcal{P}_2(\mathbb{R}^d) \to \mathbb{R}$, its first variation $\frac{\delta\mathcal{F}}{\delta\rho}$ is obtained via Gâteaux derivatives. A curve $\rho_t\in\mathcal{P}_2(\mathbb{R}^d)$ is a Wasserstein gradient flow of $\mathcal{F}$ if it satisfies the continuity equation $\partial_t\rho_t=-\nabla\cdot(\rho_t v_t)$ with velocity field $v_t=-\nabla\frac{\delta\mathcal{F}}{\delta\rho}[\rho_t]$, which can be written as the partial differential equation
\begin{equation}
\frac{\partial\rho_t}{\partial t} = \nabla\cdot\left( \rho_t \nabla\frac{\delta\mathcal{F}}{\delta\rho}[\rho_t] \right).
\end{equation}
Equivalently, the evolution can be described by the Jordan--Kinderlehrer--Otto (JKO) scheme, which recursively minimizes $\mathcal{F}[\rho]+\frac{1}{2\tau}W_2^2(\rho,\rho_t)$ over $\rho\in\mathcal{P}_2(\mathbb{R}^d)$ for a time step $\tau>0$.

\noindent \textbf{Particle Approximation.} From a computational perspective, the continuous density is approximated by an empirical measure $\rho_t^N = \frac{1}{N}\sum_{i=1}^N \delta_{X_t^{(i)}}$ represented by particles $\{X_t^{(i)}\}_{i=1}^N$, supported on $N$ particles. In this setting, the mean-field dynamics correspond to particles evolving under the velocity field:
\begin{equation}
\frac{dX_t^{(i)}}{dt} = -\nabla\frac{\delta\mathcal{F}}{\delta\rho}[\rho_t](X_t^{(i)}).
\end{equation}

This particle-based interpretation connects the macroscopic partial differential equation description with microscopic particle dynamics, thereby providing a practical framework for the numerical simulation of gradient flows. Together, these ideas form the theoretical foundation of WE, enabling the principled optimization of population distributions through gradient-flow dynamics in probability space.

\subsection{The Proposed Wasserstein Evolution Model}
To highlight the contributions of our work, this section is organized into two main components. We first examine the limitations of heuristic methods within the conventional evolutionary trajectory. This is followed by a detailed discussion of the motivation for the WE, including a description of its implementation details.

\noindent \textbf{From Heuristic Metaphors to Physical Principles.} Traditional EAs typically rely on operators such as crossover, mutation, and selection, which are inspired by biological evolution. In contrast, other approaches update solutions by constructing probabilistic models derived from population statistics, such as the mean and covariance. Representative examples of these respective paradigms include DE and CMA-ES. 

Within the DE framework, the optimization process iterates through mutation, crossover, and selection. Specifically, the mutation operator generates trial vectors as follows:
\begin{equation}
\mathbf{v}_i = \mathbf{x}_{r_1} + F \cdot (\mathbf{x}_{r_2} - \mathbf{x}_{r_3}),
\end{equation}
where $\mathbf{x}_{r_1}$, $\mathbf{x}_{r_2}$, and $\mathbf{x}_{r_3}$ denote randomly selected individuals, $F \in (0, 2]$ is the scaling factor, and $\mathbf{v}_i$ denotes the mutant vector. The binomial crossover operator then produces offspring as follows:
\begin{equation}
u_{i,j} = \begin{cases}
v_{i,j} & \text{if } \text{rand}() \leq CR \text{ or } j = j_{\text{rand}} \\
x_{i,j} & \text{otherwise}
\end{cases}
\end{equation}
where $CR \in [0, 1]$ denotes the crossover probability, and $\mathbf{u}_i$ denotes the trial vector. The greedy selection operator then computes the optimal solutions as follows:
\begin{equation}
\mathbf{x}_i^{(g+1)} = \begin{cases}
\mathbf{u}_i & \text{if } f(\mathbf{u}_i) \leq f(\mathbf{x}_i^{(g)}) \\
\mathbf{x}_i^{(g)} & \text{otherwise}
\end{cases}
\end{equation}
where $f(\cdot)$ is the objective function to be minimized, and $g$ denotes the generation index.

In contrast, within the CMA-ES framework, a multivariate Gaussian distribution $\mathcal{N}(\mathbf{m}^{(g)}, \sigma^{(g)2}\mathbf{C}^{(g)})$ is evolved, characterized by the mean $\mathbf{m}^{(g)} \in \mathbb{R}^d$, the step size $\sigma^{(g)} > 0$, and the covariance matrix $\mathbf{C}^{(g)} \in \mathbb{R}^{d \times d}$. New candidate solutions are sampled from this distribution as follows:
\begin{equation}
\mathbf{x}_k^{(g+1)} \sim \mathcal{N}(\mathbf{m}^{(g)}, \sigma^{(g)2} \mathbf{C}^{(g)}),
\end{equation}
where $k \in \{1, \dots, \lambda\}$, and $\lambda$ denotes the offspring population size. The mean is updated using weighted recombination as follows:
\begin{equation}
\mathbf{m}^{(g+1)} = \mathbf{m}^{(g)} + c_m \sigma^{(g)} \sum_{i=1}^{\mu} w_i \mathbf{y}_{i:\lambda}^{(g+1)},
\end{equation}
where $\mu \leq \lambda$ denotes the parent population size; $w_i > 0$ are the recombination weights satisfying $\sum_i w_i = 1$; $\mathbf{y}{i:\lambda}^{(g+1)} = (\mathbf{x}{i:\lambda}^{(g+1)} - \mathbf{m}^{(g)}) / \sigma^{(g)}$ denote the normalized mutation vectors; $\mathbf{x}_{i:\lambda}^{(g+1)}$ represents the $i$-th best offspring; and $c_m \in (0, 1]$ is the learning rate. The covariance matrix update combines rank-one and rank-$\mu$ contributions as follows:
\begin{equation}
\begin{aligned}
\mathbf{C}^{(g+1)} = & \ (1 - c_1 - c_\mu) \mathbf{C}^{(g)} + c_1 \mathbf{p}_c^{(g+1)} (\mathbf{p}_c^{(g+1)})^\top \\
& + c_\mu \sum_{i=1}^{\mu} w_i \mathbf{y}_{i:\lambda}^{(g+1)} (\mathbf{y}_{i:\lambda}^{(g+1)})^\top,
\end{aligned}
\end{equation}
where $\mathbf{p}c^{(g+1)} \in \mathbb{R}^d$ denotes the evolution path, and $c_1 \geq 0$ and $c_\mu \geq 0$ are learning rates satisfying $c_1 + c_\mu \leq 1$.

Despite these advances, as observed from the above equations, DE, CMA-ES, and their variants either introduce new evolutionary search operators inspired by metaphors or construct complex statistical models to update the population. This makes the regulation of exploration and exploitation challenging, as the final outcome heavily relies on Ad-hoc manual parameter settings.

This observation naturally motivates the question of \textsc{whether evolutionary algorithms can be reformulated within a physics-based framework?} in which the balance between exploration and exploitation is governed by competition among energetic driving forces represented by potential gradients.

\noindent \textbf{Derivation of Wasserstein Gradient Flow.} To response the above question, we take a step back and reconsider whether biological evolution itself can be described in physical terms. Darwinian dynamics, as a modern mathematical formulation of evolutionary mechanics, describes the evolutionary process directly through a stochastic differential equation as below:
\begin{equation}
\frac{d\mathbf{q}}{dt} = \mathbf{f}(\mathbf{q}) + \bm{\zeta}(\mathbf{q}, t),
\end{equation}
where evolution is represented by the phase-space velocity $d\mathbf{q}/dt$, the deterministic selection force by $\mathbf{f}(\mathbf{q})$, and the stochastic mutation force by $\bm{\zeta}(\mathbf{q}, t)$.

From physical perspective, changes in phase space are often accompanied by variations in macroscopic state quantities of the system, such as free energy, potential energy, and entropy. where the free energy can be expressed as:
\begin{equation}
F = U - \frac{1}{\beta} S,
\end{equation}
where $U$ represents the internal energy, $S$ denotes the entropy, and $\beta = 1/(k_B T)$ is the inverse temperature. In practical evolutionary optimization, minimization problems are typically considered, and individuals with smaller objective function values are regarded as having higher fitness. Following Wright’s argument that biological fitness corresponds to physical potential energy, we regard the optimized objective function as the potential energy $U$, while entropy is regarded as the number of microscopic states of a system. This interpretation naturally corresponds to population diversity in EAs: if the population is treated as an ensemble, individual solutions correspond to microscopic states.

During a phase transition, a system evolves from an initially random and disordered state toward a steady-state Boltzmann distribution through free energy minimization, during which the potential gradient force and the entropic force compete to drive the decrease in free energy. This physical process finds a direct analogue in evolutionary optimization: the selection force, which drives the population toward higher fitness, and the mutation force, which maintains diversity, jointly guide the population from random initialization toward optimal solutions. The intrinsic consistency between statistical physics and Darwinian dynamics motivates the proposal of WE as a physically grounded optimization framework. The core of WE is derived from the Wasserstein gradient flow of the free energy functional $\mathcal{F}[\rho]$. We define the following free-energy-optimizing objective:
\begin{equation}
\mathcal{F}[\rho] = \underbrace{\int f(\mathbf{x}) \rho(\mathbf{x}) d\mathbf{x}}_{\text{potential energy } U[\rho]} - \frac{1}{\beta} \underbrace{\left(-\int \rho(\mathbf{x}) \log \rho(\mathbf{x}) d\mathbf{x}\right)}_{\text{entropy } S[\rho]},
\end{equation}
where $f(\mathbf{x})$ denotes the objective function to be minimized, $\rho(\mathbf{x})$ represents the population density, and $\beta$ is the inverse temperature parameter controlling the exploration–exploitation trade-off. To derive the Wasserstein gradient flow, we compute the first variation of $\mathcal{F}$ with respect to $\rho$ as follows:
\begin{equation}
\begin{aligned}
\frac{\delta \mathcal{F}}{\delta \rho}[\rho](\mathbf{x}) 
&= \frac{\delta U}{\delta \rho}[\rho](\mathbf{x}) - \frac{1}{\beta} \frac{\delta S}{\delta \rho}[\rho](\mathbf{x}) \\
&= f(\mathbf{x}) - \frac{1}{\beta} \frac{\delta}{\delta \rho}\left(-\int \rho(\mathbf{x}') \log \rho(\mathbf{x}') d\mathbf{x}'\right) \\
&= f(\mathbf{x}) + \frac{1}{\beta} \left(1 + \log \rho(\mathbf{x})\right).
\end{aligned}
\end{equation}

In the context of Wasserstein gradient flows, the velocity field is given by $v_t = -\nabla \frac{\delta \mathcal{F}}{\delta \rho}[\rho_t]$. By evaluating the variational derivative, we obtain the explicit formulation:
\begin{equation}
v_t(\mathbf{x}) = -\nabla f(\mathbf{x}) - \frac{1}{\beta} \nabla \log \rho_t(\mathbf{x}),
\end{equation}
this velocity field represents the combined force acting on each point in the probability space: the first term, $-\nabla f(\mathbf{x})$, corresponds to the potential gradient force that drives particles toward lower objective values, while the second term, $-\frac{1}{\beta} \nabla \log \rho_t(\mathbf{x})$, corresponds to the entropic force that pushes particles away from high-density regions to maintain diversity. In practice, we employ an empirical distribution $\rho_t^N(\mathbf{x}) = \frac{1}{N}\sum_{i=1}^N \delta(\mathbf{x} - \mathbf{x}_i^{(t)})$, represented by particles $\{\mathbf{x}_i^{(t)}\}_{i=1}^{N}$. Substituting this empirical distribution into the particle evolution equation $\frac{d\mathbf{x}_i^{(t)}}{dt} = v_t(\mathbf{x}_i^{(t)})$ yields:
\begin{equation}
\frac{d\mathbf{x}_i^{(t)}}{dt} = -\nabla f(\mathbf{x}_i^{(t)}) - \frac{1}{\beta_t} \nabla \log \rho_t^N(\mathbf{x}_i^{(t)}),
\end{equation}
the continuous-time dynamics is further discretized using a learning rate $\eta_t$. The resulting discrete-time update rule is given as follows:
\begin{equation}
\mathbf{x}_i^{(t+1)} = \mathbf{x}_i^{(t)} + \eta_t \left[ -\nabla f(\mathbf{x}_i^{(t)}) - \frac{1}{\beta_t} \nabla \log \hat{\rho}_t(\mathbf{x}_i^{(t)}) \right],
\end{equation}
this update rule drives the population toward an equilibrium state in which the net force acting on each particle becomes zero. At equilibrium, when the population distribution stabilizes ($\mathbf{x}_i^{(t+1)} \approx \mathbf{x}_i^{(t)}$), the following condition holds:
\begin{equation}
-\nabla f(\mathbf{x}) - \frac{1}{\beta_t} \nabla \log \rho_t(\mathbf{x}) = 0.
\end{equation}

This equilibrium condition balances exploitation, driven by the potential gradient force \( -\nabla f(\mathbf{x}) \), and exploration, driven by the entropic force \( -\frac{1}{\beta_t} \nabla \log \rho_t(\mathbf{x}) \). This leads to a proportional relation between the log-density and objective gradients: \( \nabla \log \rho_t(\mathbf{x}) = -\beta_t \nabla f(\mathbf{x}) \). Integrating and exponentiating yields the unnormalized Boltzmann form: \( \rho_t(\mathbf{x}) \propto e^{-\beta_t f(\mathbf{x})} \), establishing a direct link between the population density and the objective function.

To obtain a valid probability density function that integrates to one, we normalize the distribution by introducing the partition function $Z_{\beta_t}$:
\begin{equation}
\rho_{\text{eq}}(\mathbf{x}) = \frac{1}{Z_{\beta_t}} e^{-\beta_t f(\mathbf{x})}, \quad \text{where} \quad Z_{\beta_t} = \int_{\mathcal{X}} e^{-\beta_t f(\mathbf{x})} d\mathbf{x},
\end{equation}
the above expression corresponds precisely to the Boltzmann distribution in statistical physics, with $\beta_t$ acting as an inverse temperature parameter. The distribution has several important properties: (1) It assigns higher probability to regions with lower function values, ensuring convergence to high-quality solutions. (2) It maintains non-zero probability over the entire search space, preserving diversity. (3) As $\beta_t \to \infty$ (analogous to temperature approaching zero), the distribution concentrates on the global minima of $f(\mathbf{x})$. This theoretical convergence guarantee distinguishes WE from traditional EAs and provides a principled foundation for balancing exploration and exploitation through free energy minimization.The pseudo-code of WE implementing this update is presented in Algorithm \ref{alg:algorithm1}, and its overall framework is illustrated in Figure ~\ref{fig:fra}. Appendix A.3 provides the convergence analysis of the WE.

\begin{figure*}[t]
    \centering
    \includegraphics[width=\linewidth]{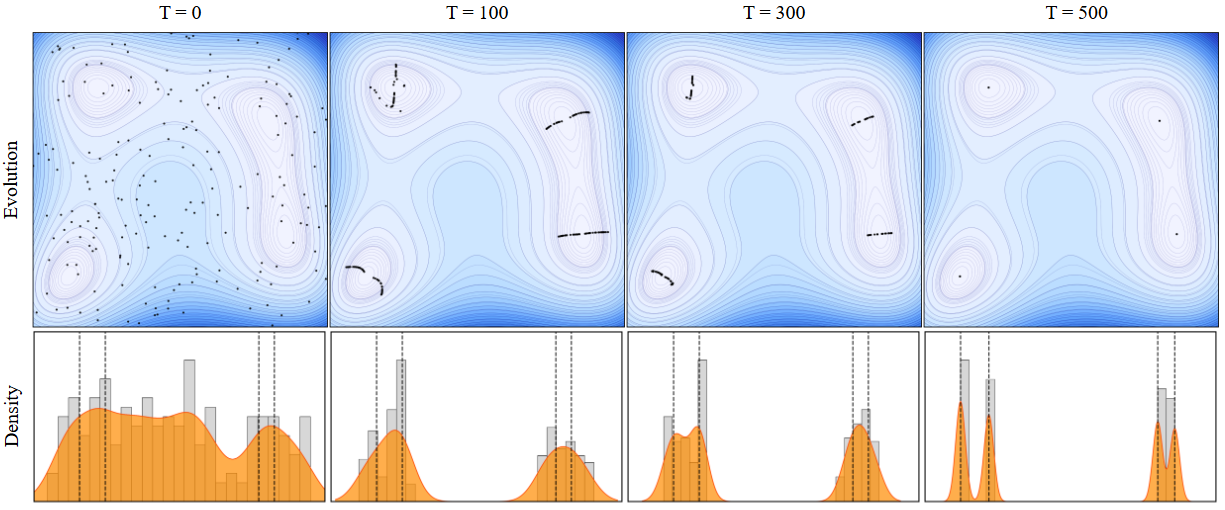}
    \caption{Evolution of the WE population distribution on the Himmelblau function. (Top) Spatial distribution of solutions (black dots) at iterations. (Bottom) Marginal frequency distribution along the first dimension with corresponding KDE curves. Dashed lines indicate the four theoretical minima, while the orange shading represents the reference probability density function. The population evolves from a random initialization toward a Boltzmann distribution, successfully covering all four global optima.
}
    \label{fig:bar}
\end{figure*}

\begin{algorithm}[tb]
    \caption{Wasserstein Evolution (WE)}
    \label{alg:algorithm1}
    \begin{algorithmic}[1]
        \REQUIRE Fitness function $f: \Omega \to \mathbb{R}$, Population size $N$, Max generations $T$, Learning rate schedule $\{\eta_t\}$.
        \ENSURE Best solution $\mathbf{x}^*$ and fitness $f^*$.
        
        \STATE Initialize population $\{\mathbf{x}_i^{(0)}\}_{i=1}^N \sim \text{Uniform}(\Omega)$
        \FOR{$t = 0, 1, \dots, T-1$}
            \STATE Update inverse temperature $\beta_t$
            \STATE Estimate density $\hat{\rho}_t(\mathbf{x})$ using kernel density estimation
            
            \FOR{$i = 1$ to $N$}
                \STATE Calculate gradients:
                \STATE \quad $\mathbf{g}_{\text{fit}} \leftarrow \nabla f(\mathbf{x}_i^{(t)})$
                \STATE \quad $\mathbf{g}_{\text{ent}} \leftarrow \nabla \log \hat{\rho}_t(\mathbf{x}_i^{(t)})$
                \STATE Update position:
                \STATE \quad $\mathbf{x}_i^{(t+1)} \leftarrow \mathbf{x}_i^{(t)} - \eta_t \left(\mathbf{g}_{\text{fit}} + \beta_t^{-1} \mathbf{g}_{\text{ent}}\right)$
            \ENDFOR
            
            \STATE Track current best: $\mathbf{x}^* \leftarrow \arg\min_{\mathbf{x} \in \{\mathbf{x}_i^{(t+1)}\}} f(\mathbf{x})$
        \ENDFOR
        \STATE \textbf{return} $\mathbf{x}^*, f(\mathbf{x}^*)$
    \end{algorithmic}
\end{algorithm}
\begin{figure}[t]
	\centering
	\includegraphics[width=\linewidth]{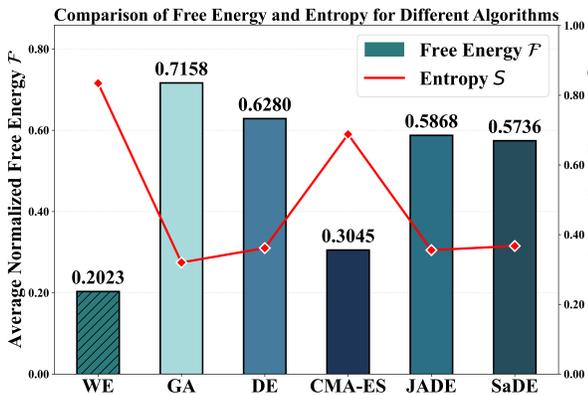}
	\caption{Comparison of thermodynamic metrics. We plot Average Entropy (lines) and Average Free Energy (bars) for WE against five evolutionary baselines. WE consistently demonstrates higher diversity and lower free energy, indicating superior stability compared to competitors.}
	\label{fig:FreeEnergy_Entropy}
\end{figure}
\section{Experiments}

\subsection{Experimental Setup}

We conducted comprehensive experiments to evaluate WE against five established EAs: GA, DE, CMA-ES, JADE and SaDE. We set the population size to (N = 30) and the number of generations to (T = 500) for each algorithm. All results are averaged over 30 independent runs. Higher entropy corresponds to a larger number of accessible microscopic states, reflecting greater population diversity. In addition, to examine whether the Wasserstein gradient flow framework enables WE to achieve lower free energy, we compute the average free energy. From a physical perspective, lower free energy represents a more stable state of the system. The experiments were implemented using Python 3.9 and the PyTorch framework.Appendix A.1 and A.2 detail the specific forms of the test functions required. The computational details for the entropy $S$ and the free energy $\mathcal{F}$ are provided in Appendix~A.4.

\subsection{Experiments on Benchmark Functions}
\begin{table}[t]
    \caption{Quantitative comparison of population diversity and convergence stability. We report Average Entropy $S$ ($\uparrow$) and Free Energy $\mathcal{F}$ ($\downarrow$) on benchmark functions. $\mathcal{F}$ reflects the stability achieved via Wasserstein gradient flow. \textbf{Bold} values with $^{*}$ denote statistically significant improvements over all baselines (Wilcoxon test, $p<0.05$).}
    \centering
    \scriptsize
    \renewcommand{\arraystretch}{1.1}
    \setlength{\tabcolsep}{1.5pt}
    \begin{tabular}{lccccccc}
        \hline
        \textbf{Function (Metric)} & \textbf{WE} & \textbf{GA} & \textbf{DE} & \textbf{CMA-ES} & \textbf{JADE} & \textbf{SaDE} \\
        \hline
        \textbf{Rastrigin} ($S \uparrow$) & \textbf{0.9034}$^{*}$ & 0.3895 & 0.3907 & 0.3913 & 0.3897 & 0.3889 \\
        \textbf{Rastrigin} ($\mathcal{F} \downarrow$) & \textbf{0.1451}$^{*}$ & 0.8441 & 0.5053 & 0.5041 & 0.5053 & 0.5065 \\
        \hline
        \textbf{Beale} ($S \uparrow$) & \textbf{0.9033}$^{*}$ & 0.4091 & 0.3854 & 0.3858 & 0.3850 & 0.3854 \\
        \textbf{Beale} ($\mathcal{F} \downarrow$) & \textbf{0.3338}$^{*}$ & 0.9022 & 0.4055 & 0.4055 & 0.4055 & 0.4055 \\
        \hline
        \textbf{Himmelblau} ($S \uparrow$) & \textbf{0.8456}$^{*}$ & 0.3321 & 0.3330 & 0.7470 & 0.3328 & 0.3331 \\
        \textbf{Himmelblau} ($\mathcal{F} \downarrow$) & \textbf{0.1468}$^{*}$ & 0.7246 & 0.6426 & 0.2754 & 0.6426 & 0.6426 \\
        \hline
        \textbf{Six-Hump Camel} ($S \uparrow$) & \textbf{0.8324}$^{*}$ & 0.3289 & 0.3298 & 0.7697 & 0.3372 & 0.3292 \\
        \textbf{Six-Hump Camel} ($\mathcal{F} \downarrow$) & \textbf{0.1678}$^{*}$ & 0.6730 & 0.6702 & 0.2299 & 0.6616 & 0.6702 \\
        \hline
        \textbf{Holder Table} ($S \uparrow$) & \textbf{0.8248}$^{*}$ & 0.3170 & 0.3375 & 0.7802 & 0.3472 & 0.3224 \\
        \textbf{Holder Table} ($\mathcal{F} \downarrow$) & \textbf{0.2677}$^{*}$ & 0.4364 & 0.8963 & 0.4315 & 0.4191 & 0.4275 \\
        \hline
        \textbf{2D Periodic} ($S \uparrow$) & \textbf{0.7465}$^{*}$ & 0.1288 & 0.4971 & 0.6879 & 0.3777 & 0.6346 \\
        \textbf{2D Periodic} ($\mathcal{F} \downarrow$) & \textbf{0.2535}$^{*}$ & 0.8714 & 0.5027 & 0.3131 & 0.6216 & 0.3648 \\
        \hline
        \textbf{Double-Well} ($S \uparrow$) & \textbf{0.8126}$^{*}$ & 0.3128 & 0.3397 & 0.7945 & 0.3589 & 0.3126 \\
        \textbf{Double-Well} ($\mathcal{F} \downarrow$) & \textbf{0.1873}$^{*}$ & 0.6872 & 0.6601 & 0.2056 & 0.6415 & 0.6872 \\
        \hline
        \textbf{Tokamak} ($S \uparrow$) & \textbf{0.8174}$^{*}$ & 0.3218 & 0.3359 & 0.7900 & 0.3366 & 0.3277 \\
        \textbf{Tokamak} ($\mathcal{F} \downarrow$) & \textbf{0.1824}$^{*}$ & 0.6782 & 0.6634 & 0.2102 & 0.6634 & 0.6729 \\
        \hline
        \textbf{Multipole} ($S \uparrow$) & \textbf{0.8534}$^{*}$ & 0.3307 & 0.3463 & 0.7301 & 0.3390 & 0.3211 \\
        \textbf{Multipole} ($\mathcal{F} \downarrow$) & \textbf{0.1465}$^{*}$ & 0.6709 & 0.6532 & 0.2701 & 0.6606 & 0.6781 \\
        \hline
        \textbf{Optical Lattice} ($S \uparrow$) & \textbf{0.8047}$^{*}$ & 0.3307 & 0.3198 & 0.8037 & 0.3522 & 0.3194 \\
        \textbf{Optical Lattice} ($\mathcal{F} \downarrow$) & \textbf{0.1923}$^{*}$ & 0.6702 & 0.6802 & 0.1995 & 0.6472 & 0.6802 \\
        \hline
    \end{tabular}
    \label{tab:all_experimental_results_normalized}
\end{table}

\begin{figure}[t]
    \centering
    \includegraphics[width=\linewidth]{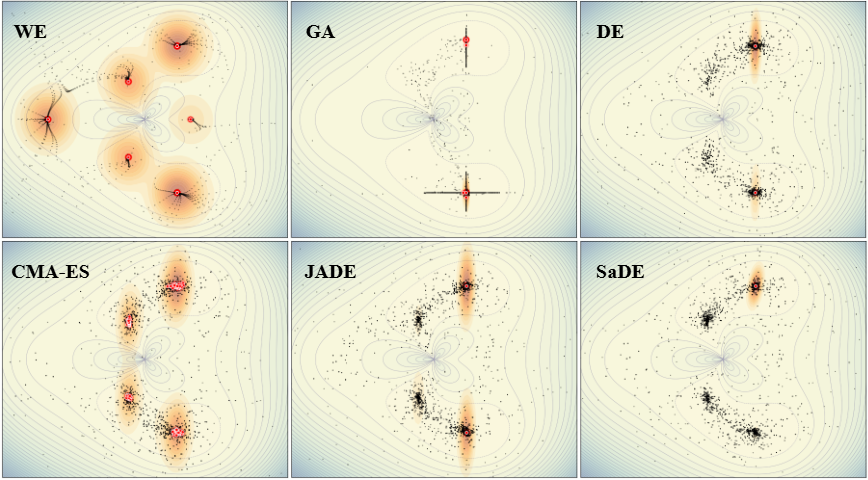}
    \caption{Trajectory comparison on the Tokamak function. Black dots trace the historical search paths, while orange dots and halos represent the final population and its probability density, respectively. WE exhibits structured convergence, successfully discovering multiple modes and centering the final distribution around distinct local minima.}
    \label{fig:Trajectory}
\end{figure}

The performance of all six algorithms across the ten benchmark functions is summarized in Table \ref{tab:all_experimental_results_normalized}, covering five classical multimodal functions and five real-world physical potential functions. Each cell reports the average  Entropy $S$ and the corresponding average Free Energy $\mathcal{F}$ over 30 independent runs.Figure~\ref{fig:bar} displays the evolution of the population distribution throughout the WE iterations, and Figure~\ref{fig:Trajectory} illustrates the population trajectories and final probability distributions of different algorithms during the iterative process.Figure~\ref{fig:FreeEnergy_Entropy} visually demonstrates the comparison of normalized Free Energy (bars) and Entropy (line) across all six algorithms. The results clearly show that the proposed WE consistently achieves the highest Entropy while maintaining the lowest Free Energy.

\begin{table}[t]
    \caption{Quantitative results on Schwefel 2.22 Transformations. We report Average Entropy $S$ ($\uparrow$) and Free Energy $\mathcal{F}$ ($\downarrow$). Higher $S$ denotes better diversity, while lower $\mathcal{F}$ indicates stability achieved via Wasserstein gradient flow. \textbf{Bold} values with $^{*}$ indicate statistically significant improvements (Wilcoxon test, $p<0.05$).}
    \centering
    \scriptsize
    \renewcommand{\arraystretch}{1.1}
    \setlength{\tabcolsep}{1.5pt}
    \begin{tabular}{lccccccc}
        \hline
        \textbf{Function (Metric)} & \textbf{WE} & \textbf{GA} & \textbf{DE} & \textbf{CMA-ES} & \textbf{JADE} & \textbf{SaDE} \\
        \hline
        \textbf{Original} ($S \uparrow$) & \textbf{0.9034}$^{*}$ & 0.4009 & 0.3875 & 0.3871 & 0.3871 & 0.3878 \\
        \textbf{Original} ($\mathcal{F} \downarrow$) & \textbf{0.3596}$^{*}$ & 0.9031 & 0.3980 & 0.3981 & 0.3981 & 0.3980 \\
        \hline
        \textbf{Shift Right 20} ($S \uparrow$) & \textbf{0.9033}$^{*}$ & 0.4070 & 0.3857 & 0.3849 & 0.3865 & 0.3865 \\
        \textbf{Shift Right 20} ($\mathcal{F} \downarrow$) & \textbf{0.3695}$^{*}$ & 0.9033 & 0.3953 & 0.3954 & 0.3953 & 0.3953 \\
        \hline
        \textbf{Shift Left 30} ($S \uparrow$) & \textbf{0.9034}$^{*}$ & 0.3988 & 0.3879 & 0.3876 & 0.3878 & 0.3882 \\
        \textbf{Shift Left 30} ($\mathcal{F} \downarrow$) & \textbf{0.3683}$^{*}$ & 0.9033 & 0.3956 & 0.3956 & 0.3956 & 0.3956 \\
        \hline
        \textbf{Shift (15, 15)} ($S \uparrow$) & \textbf{0.9033}$^{*}$ & 0.4090 & 0.3858 & 0.3847 & 0.3857 & 0.3854 \\
        \textbf{Shift (15, 15)} ($\mathcal{F} \downarrow$) & \textbf{0.3584}$^{*}$ & 0.9031 & 0.3983 & 0.3985 & 0.3983 & 0.3983 \\
        \hline
        \textbf{Scale x2} ($S \uparrow$) & \textbf{0.9034}$^{*}$ & 0.3978 & 0.3885 & 0.3883 & 0.3875 & 0.3882 \\
        \textbf{Scale x2} ($\mathcal{F} \downarrow$) & \textbf{0.2832}$^{*}$ & 0.8983 & 0.4225 & 0.4225 & 0.4225 & 0.4225 \\
        \hline
        \textbf{Scale x0.5} ($S \uparrow$) & \textbf{0.9034}$^{*}$ & 0.4015 & 0.3883 & 0.3872 & 0.3868 & 0.3865 \\
        \textbf{Scale x0.5} ($\mathcal{F} \downarrow$) & \textbf{0.3830}$^{*}$ & 0.9034 & 0.3918 & 0.3918 & 0.3918 & 0.3918 \\
        \hline
        \textbf{Anisotropic Scale} ($S \uparrow$) & \textbf{0.9034}$^{*}$ & 0.4008 & 0.3872 & 0.3881 & 0.3861 & 0.3881 \\
        \textbf{Anisotropic Scale} ($\mathcal{F} \downarrow$) & \textbf{0.3392}$^{*}$ & 0.9024 & 0.4038 & 0.4038 & 0.4040 & 0.4038 \\
        \hline
        \textbf{Rotate 45°} ($S \uparrow$) & \textbf{0.9033}$^{*}$ & 0.4117 & 0.3848 & 0.3847 & 0.3852 & 0.3845 \\
        \textbf{Rotate 45°} ($\mathcal{F} \downarrow$) & \textbf{0.3207}$^{*}$ & 0.9015 & 0.4094 & 0.4097 & 0.4094 & 0.4097 \\
        \hline
        \textbf{Rotate -30°} ($S \uparrow$) & \textbf{0.9034}$^{*}$ & 0.3889 & 0.3903 & 0.3902 & 0.3902 & 0.3905 \\
        \textbf{Rotate -30°} ($\mathcal{F} \downarrow$) & \textbf{0.3637}$^{*}$ & 0.9032 & 0.3969 & 0.3969 & 0.3969 & 0.3969 \\
        \hline
        \textbf{Rotate 75°} ($S \uparrow$) & \textbf{0.9034}$^{*}$ & 0.3904 & 0.3896 & 0.3905 & 0.3903 & 0.3893 \\
        \textbf{Rotate 75°} ($\mathcal{F} \downarrow$) & \textbf{0.3548}$^{*}$ & 0.9030 & 0.3994 & 0.3993 & 0.3993 & 0.3994 \\
        \hline
        \textbf{Translate-Rotate-Scale} ($S \uparrow$) & \textbf{0.9034}$^{*}$ & 0.3899 & 0.3906 & 0.3905 & 0.3887 & 0.3904 \\
        \textbf{Translate-Rotate-Scale} ($\mathcal{F} \downarrow$) & \textbf{0.3306}$^{*}$ & 0.9020 & 0.4064 & 0.4064 & 0.4067 & 0.4064 \\
        \hline
        \textbf{Rotate-Translate-Scale} ($S \uparrow$) & \textbf{0.9034}$^{*}$ & 0.3888 & 0.3899 & 0.3900 & 0.3904 & 0.3910 \\
        \textbf{Rotate-Translate-Scale} ($\mathcal{F} \downarrow$) & \textbf{0.3186}$^{*}$ & 0.9014 & 0.4102 & 0.4102 & 0.4102 & 0.4102 \\
        \hline
    \end{tabular}
    \label{tab:invariance_experimental_results_normalized}
\end{table}

\begin{figure}[t]
	\centering
	\includegraphics[width=\linewidth]{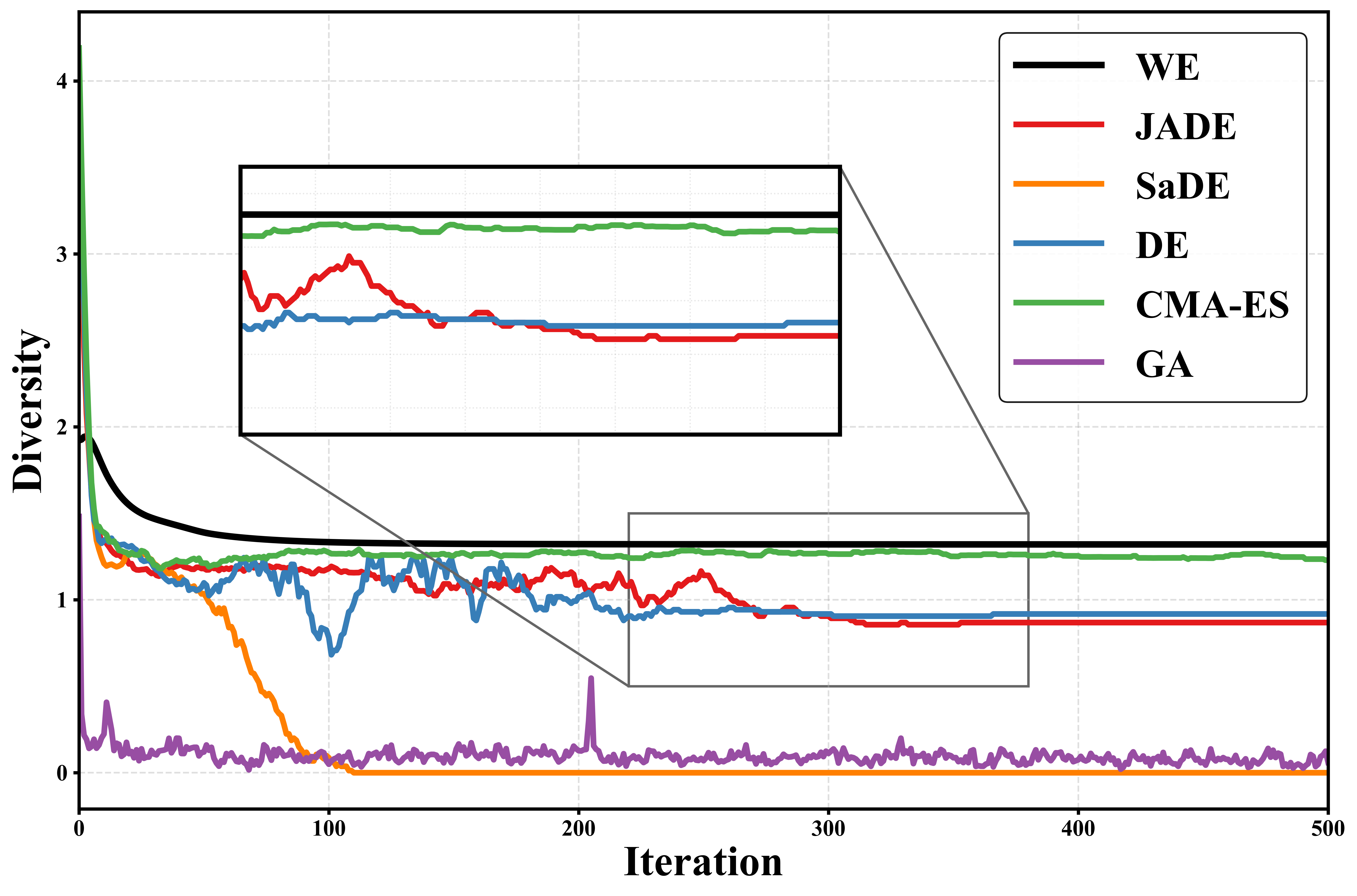}
	\caption{Evolution of population diversity over iterations. This visualization corroborates the entropy analysis (methodology detailed in Appendix A.4). WE consistently sustains the highest diversity levels during late-stage evolution, significantly outperforming all competing algorithms.}
	\label{fig:diversity}
\end{figure}

Across all benchmark functions, WE consistently achieves the highest entropy and the lowest average free energy. Wilcoxon signed-rank tests further confirm that the differences between WE and all competing algorithms are statistically significant, thereby demonstrating its clear superiority. Among the baseline methods, GA exhibits the highest free energy. Although GA preserves entropy at a level comparable to DE, JADE, and SaDE, its limited convergence leads to suboptimal energy performance. The DE variants (DE, JADE, and SaDE) exhibit largely similar performance across most benchmarks, with noticeable divergence only on the real-world 2D Periodic function. Nevertheless, all are substantially outperformed by WE. CMA-ES achieves the second-best overall performance by effectively balancing convergence and population diversity, a characteristic of evolution strategy–based approaches.As a supplementary illustration, Figure~\ref{fig:diversity} shows the evolution of population diversity across iterations for different algorithms on the Tokamak function.

\subsection{Robustness of WE in Different Invariance}
Invariance to transformations of the fitness landscape, including translation, rotation, and scaling, is a critical property of effective EAs \cite{kudela2022critical,tian2023principled}. To assess this property, the proposed WE is evaluated on the Schwefel 2.22 benchmark and its transformed variants, which are described in Appendix A.2. Table \ref{tab:invariance_experimental_results_normalized} presents comparative results in terms of entropy and free energy, averaged over 30 independent runs, and contrasts WE with established baseline algorithms.

Experimental results indicate that WE consistently achieves the highest entropy and the lowest average free energy on both the original Schwefel 2.22 function and its transformed variants, with statistically significant improvements over all baseline algorithms. Among the competing algorithms, GA shows the weakest performance and exhibits the highest free energy due to its limited convergence capability. In contrast, DE, CMAES, JADE, and SaDE demonstrate broadly comparable but still suboptimal performance, and all consistently lag behind WE. These findings confirm the robustness of WE to transformations of the fitness landscape, including translation, rotation, and scaling, as well as its superior ability to balance convergence and population diversity. At a more fundamental level, the results validate the physical mechanism underlying WE. By explicitly minimizing free energy as an intrinsic objective, the algorithm theoretically converges toward the Boltzmann distribution and ensures the maintenance of high entropy throughout the search process.

\section{Conclusion}
In conclusion, this work establishes a connection between evolutionary optimization and statistical physics from the perspective of phase transition theory. By rigorously formulating the convergence dynamics of evolutionary algorithms as a free energy minimization process governed by Wasserstein gradient flows, this study proposes Wasserstein Evolution, referred to as WE. Unlike traditional approaches, WE adaptively regulates the balance between exploration and exploitation by jointly considering potential energy and entropy, which theoretically guarantees convergence to the Boltzmann distribution. This mechanism, grounded in physical principles, enhances both interpretability and robustness. Extensive experiments involving five widely used baseline algorithms further demonstrate that WE consistently achieves the lowest free energy and the highest entropy, which validates its overall superiority. Future work will focus on integrating the WE into broader stochastic optimization frameworks to address complex practical applications, such as neural network training and engineering design optimization.

\section*{Ethical Statement}

There are no ethical issues.

\bibliographystyle{named}
\bibliography{ijcai26}

\end{document}



\section*{Appendix}
\subsection*{A.1 Benchmark Functions}
This subsection details the benchmark functions used to evaluate the proposed algorithm and its competitors.The experimental suite comprises ten carefully selected functions, which can be broadly divided into two categories: five classical, well-established multimodal functions that present complex search landscapes with multiple local and global optima, and five real-world physical potential functions that model actual phenomena from physics and engineering. The latter category tests the algorithm's capability to navigate complex, physically meaningful energy landscapes. The mathematical expressions, search domains, and key characteristics of each function are provided in Table~\ref{tab:all_test_functions}.
\begin{table*}[t]
  \caption{Benchmark and Physical Potential Functions Used in the Experiments}
  \centering
  \scriptsize
  \setlength{\tabcolsep}{4pt}
  \renewcommand{\arraystretch}{1.15}

  \begin{tabularx}{\textwidth}{@{}>{\raggedright\arraybackslash}X c >{\raggedright\arraybackslash}X@{}}
    \toprule
    \textbf{Name \& Formula} & \textbf{Domain} & \textbf{Features / Physical Interpretation} \\
    \midrule
    \textbf{Rastrigin}: $f(\mathbf{x}) = 10n + \sum_{i=1}^{n} \left[ x_i^2 - 10\cos(2\pi x_i) \right]$
    & $[-5.12, 5.12]^2$
    & Highly multimodal, separable. Global optimum: $f(\mathbf{0}) = 0$. \\
    \addlinespace[3pt]
    \textbf{Beale}: $f(x, y) = (1.5 - x + xy)^2 + (2.25 - x + xy^2)^2 + (2.625 - x + xy^3)^2$
    & $[-4.5, 4.5]^2$
    & Non-convex, unimodal with a narrow valley. Global optimum: $f(3, 0.5) = 0$. \\
    \addlinespace[3pt]
    \textbf{Himmelblau}: $f(x, y) = (x^2 + y - 11)^2 + (x + y^2 - 7)^2$
    & $[-6, 6]^2$
    & Symmetric with four equal-valued global minima ($f^*=0$). \\
    \addlinespace[3pt]
    \textbf{Six-Hump Camel}: $f(x, y) = \left(4 - 2.1x^2 + \frac{x^4}{3}\right)x^2 + xy + (-4 + 4y^2)y^2$
    & $x\in[-3,3], y\in[-2,2]$
    & Two global minima ($f\approx -1.0316$) and four local minima. \\
    \addlinespace[3pt]
    \textbf{Holder Table}: $f(x, y) = -\left| \sin(x) \cos(y) \exp\left( \left| 1 - \frac{\sqrt{x^2 + y^2}}{\pi} \right| \right) \right|$
    & $[-10, 10]^2$
    & Highly multimodal. Four global minima ($f\approx -19.2085$). \\
    \addlinespace[5pt]
    \textbf{2D Periodic}: $V = -V_0[\cos(\frac{2\pi x}{a}) + \cos(\frac{2\pi y}{a}) + \lambda_c\cos(\frac{2\pi x}{a})\cos(\frac{2\pi y}{a})], V_0=2.0, a=1.0, \lambda_c=0.5$
    & $[-2, 2]^2$
    & Periodic potential for crystal adsorption. Symmetric lattice wells. \\
    \addlinespace[3pt]
    \textbf{Double-Well}: $V = (x^2 - 1)^2 + \frac{1}{2}ky^2 + A_g\exp[-B(x^2 + y^2)], k=2.0, A_g=3.0, B=3.0$
    & $[-2, 2]^2$
    & Molecular conformation transitions. Two metastable states separated by a Gaussian barrier. \\
    \addlinespace[3pt]
    \textbf{Tokamak}: $V = \alpha(r - r_0)^2 + \beta(r - r_0)^4 + \epsilon(r - r_0)\cos(m\theta) + \delta\cos(n\theta), r_0=2.0, \alpha=0.5, \beta=0.05, \epsilon=0.8, \delta=0.6, m=3, n=2$
    & $[-4, 4]^2$
    & Charged particle drift in tokamak fields. Magnetic islands create multiple metastable states. \\
    \addlinespace[3pt]
    \textbf{Multipole}: $V = \sum_{i=1}^{4}\frac{s_i q}{4\pi\epsilon_0 \sqrt{(x-x_i)^2 + (y-y_i)^2 + \delta^2}}, q=1.0, \epsilon_0=1.0, \delta=0.3$
    & $[-3, 3]^2$
    & Quadrupole ion trap potential. Alternating charges create saddle points and trapping regions. \\
    \addlinespace[3pt]
    \textbf{Optical Lattice}: $V = V_0[\sin^2(kx) + \sin^2(ky)] \exp[-\frac{x^2+y^2}{2\sigma^2}], V_0=3.0, k=2.0, \sigma=3.0$
    & $[-4, 4]^2$
    & Optical lattice for cold atoms. Periodic modulation with Gaussian envelope creates multiple traps. \\
    \bottomrule
  \end{tabularx}

  \label{tab:all_test_functions}
\end{table*}

\subsection*{A.2 Schwefel 2.22 Function and Its Transformations}

To systematically evaluate the robustness and invariance properties of Wasserstein Evolution (WE), we employ the Schwefel 2.22 function as a base test case. We generate a comprehensive suite of variants by applying different levels of translation, scaling, rotation, and composite transformations to the original function. This allows us to rigorously test WE's ability to maintain stable convergence when the fitness landscape undergoes systematic perturbations, thereby assessing its invariance to fundamental geometric changes. All transformation details are summarized in Table~\ref{tab:schwefel_transforms}.
\begin{table*}[t]
  \caption{Schwefel 2.22 Function and Its Transformations for Invariance Testing}
  \centering
  \scriptsize
  \setlength{\tabcolsep}{5pt}
  \renewcommand{\arraystretch}{1.15}

  \begin{tabularx}{\textwidth}{@{}L{0.12\textwidth} L{0.26\textwidth} c X@{}}
    \toprule
    \textbf{Category} & \textbf{Name \& Description} & \textbf{Domain} & \textbf{Transformation Form / Notes} \\
    \midrule

    \textbf{Base Function} &
    \textbf{Original Schwefel 2.22} &
    $[-100, 100]^2$ &
    $f(\mathbf{x}) = \sum_{i=1}^{n} \left( \sum_{j=1}^{i} x_j \right)^2$, $n=2$. Global minimum: $f(\mathbf{0}) = 0$. \\
    \midrule

    \textbf{Translation} &
    \textbf{Shift Right 20} &
    $[-100, 100]^2$ &
    $\mathbf{x}' = \mathbf{x} - (20,20)$. Minimum at $\mathbf{x}^* = (20, 20)$. \\
    \addlinespace[2pt]
    &
    \textbf{Shift Left 30} &
    $[-100, 100]^2$ &
    $\mathbf{x}' = \mathbf{x} + (30,30)$. Minimum at $\mathbf{x}^* = (-30, -30)$. \\
    \addlinespace[2pt]
    &
    \textbf{Shift (15, 15)} &
    $[-100, 100]^2$ &
    $\mathbf{x}' = \mathbf{x} - (15,15)$. Minimum at $\mathbf{x}^* = (15, 15)$. \\
    \midrule

    \textbf{Scaling} &
    \textbf{Scale ×2 (Enlarge)} &
    $[-100, 100]^2$ &
    $\mathbf{x}' = \mathbf{x} / 2$. Landscape enlarged by factor 2. \\
    \addlinespace[2pt]
    &
    \textbf{Scale ×0.5 (Reduce)} &
    $[-100, 100]^2$ &
    $\mathbf{x}' = 2\mathbf{x}$. Landscape shrunk by factor 0.5. \\
    \addlinespace[2pt]
    &
    \textbf{Anisotropic Scale} &
    $[-100, 100]^2$ &
    $x_1' = 0.6667x_1$, $x_2' = 1.25x_2$. Different scaling per axis. \\
    \midrule

    \textbf{Rotation} &
    \textbf{Rotate 45°} &
    $[-100, 100]^2$ &
    $\mathbf{x}' = R(45^\circ)\mathbf{x}$, where $R$ is a 2D rotation matrix. \\
    \addlinespace[2pt]
    &
    \textbf{Rotate -30°} &
    $[-100, 100]^2$ &
    $\mathbf{x}' = R(-30^\circ)\mathbf{x}$. \\
    \addlinespace[2pt]
    &
    \textbf{Rotate 75°} &
    $[-100, 100]^2$ &
    $\mathbf{x}' = R(75^\circ)\mathbf{x}$. \\
    \midrule

    \textbf{Composite} &
    \textbf{Translate-Rotate-Scale} &
    $[-100, 100]^2$ &
    Sequence: $\mathbf{x} \rightarrow \mathbf{x} - (10,-10) \rightarrow R(60^\circ)\mathbf{x} \rightarrow (0.8x_1, 1.2x_2)$. \\
    \addlinespace[2pt]
    &
    \textbf{Rotate-Translate-Scale} &
    $[-100, 100]^2$ &
    Sequence: $\mathbf{x} \rightarrow R(45^\circ)\mathbf{x} \rightarrow \mathbf{x} + (15,-15) \rightarrow (0.7x_1, 1.5x_2)$. \\
    \bottomrule
  \end{tabularx}

  \label{tab:schwefel_transforms}
\end{table*}

\subsection*{A.3 Convergence Analysis of the Wasserstein Evolution}

We provide a comprehensive convergence analysis of the Wasserstein Evolution (WE). The algorithm operates on a search space $\mathcal{X} \subseteq \mathbb{R}^d$ with objective function $f: \mathcal{X} \to \mathbb{R}$. We consider the minimization problem where we aim to find $\min_{x \in \mathcal{X}} f(x)$. At each iteration $t$, the algorithm maintains $N$ particles $X_t^{(1)}, X_t^{(2)}, \dots, X_t^{(N)}$ that evolve according to deterministic dynamics.

The core update equation for each particle is given by:
\begin{equation}
	\frac{dX^{(i)}}{dt} = -\nabla f(X^{(i)}) - \frac{1}{\beta_t}\nabla\log\rho_t(X^{(i)}),
	\label{eq:particle_update}
\end{equation}
where $\rho_t(x)$ represents the probability density of the particle distribution at time $t$, and $\beta_t > 0$ is the inverse temperature parameter that varies with time. The first term $-\nabla f(X^{(i)})$ drives particles toward regions of lower function values (for minimization problems), while the second term $-\frac{1}{\beta_t}\nabla\log\rho_t(X^{(i)})$ acts as a repulsive entropic force that prevents particle concentration and maintains diversity.

Rather than analyzing individual particle trajectories, we consider the evolution of the entire probability distribution $\rho_t(x)$. The empirical distribution formed by the particles approximates this continuous density. The relationship between particle motion and distribution evolution is governed by the continuity equation, which states that the change in probability density equals the negative divergence of the probability flux. Substituting the velocity field from equation \eqref{eq:particle_update} yields the partial differential equation:
\begin{equation}
	\frac{\partial \rho_t(x)}{\partial t} = -\nabla \cdot \left[\rho_t(x)\left(-\nabla f(x) - \frac{1}{\beta_t}\nabla\log\rho_t(x)\right)\right].
	\label{eq:continuity}
\end{equation}
This equation describes how the probability distribution evolves over time in response to the combined forces of potential gradient and entropic repulsion.

To analyze convergence, we introduce the free energy functional defined as:
\begin{equation}
	\mathcal{F}_{\beta_t}[\rho] = \mathbb{E}_{x\sim\rho}[f(x)] - \frac{1}{\beta_t}S[\rho],
	\label{eq:free_energy}
\end{equation}
where $S[\rho] = -\int \rho(x)\log\rho(x) dx$ is the entropy. The first term $\mathbb{E}_{x\sim\rho}[f(x)] = \int f(x)\rho(x) dx$ represents the average objective function value, while the second term $-\frac{1}{\beta_t}S[\rho]$ represents the entropic contribution.

We now compute the time derivative of the free energy along the solution trajectory of equation \eqref{eq:continuity}. First, we compute the functional derivative:
\begin{equation}
	\frac{\delta\mathcal{F}_{\beta_t}[\rho]}{\delta\rho(x)} = f(x) + \frac{1}{\beta_t}(\log\rho(x) + 1).
	\label{eq:functional_derivative}
\end{equation}

The time derivative is then:
\begin{equation}
	\frac{d}{dt}\mathcal{F}_{\beta_t}[\rho_t] = \int_{\mathcal{X}} \frac{\delta\mathcal{F}_{\beta_t}[\rho_t]}{\delta\rho(x)} \frac{\partial\rho_t(x)}{\partial t} dx.
\end{equation}

Substituting equation \eqref{eq:continuity} gives:

\begin{equation}
\begin{aligned}
	\frac{d}{dt}\mathcal{F}_{\beta_t}[\rho_t] &= \int_{\mathcal{X}} \Big[f(x) + \frac{1}{\beta_t}(\log\rho_t(x) + 1)\Big] \\
	&\quad\times \nabla \cdot \Big[\rho_t(x)\Big(\nabla f(x) + \frac{1}{\beta_t}\nabla\log\rho_t(x)\Big)\Big] dx.
\end{aligned}
\end{equation}

Applying integration by parts with the assumption that boundary terms vanish (which holds for compact domains or rapidly decaying distributions), we obtain:
\begin{equation}
\begin{aligned}
	\frac{d}{dt}\mathcal{F}_{\beta_t}[\rho_t] &= \int_{\mathcal{X}} \nabla\left[f(x) + \frac{1}{\beta_t}(\log\rho_t(x) + 1)\right] \\
	&\quad\cdot \left[\rho_t(x)\left(-\nabla f(x) - \frac{1}{\beta_t}\nabla\log\rho_t(x)\right)\right] dx.
\end{aligned}
\end{equation}
Evaluating the gradient inside the integral yields:

\begin{equation}
\begin{aligned}
	\frac{d}{dt}\mathcal{F}_{\beta_t}[\rho_t] &= \int_{\mathcal{X}} \rho_t(x) \left(\nabla f(x) + \frac{1}{\beta_t}\nabla\log\rho_t(x)\right) \\
	&\quad\cdot \left(-\nabla f(x) - \frac{1}{\beta_t}\nabla\log\rho_t(x)\right) dx.
\end{aligned}
\end{equation}

This simplifies to:
\begin{equation}
	\frac{d}{dt}\mathcal{F}_{\beta_t}[\rho_t] = -\int_{\mathcal{X}} \rho_t(x) \left\|\nabla f(x) + \frac{1}{\beta_t}\nabla\log\rho_t(x)\right\|^2 dx \leq 0.
	\label{eq:free_energy_dissipation}
\end{equation}

Equation \eqref{eq:free_energy_dissipation} shows that the free energy is non-increasing along the trajectory, establishing $\mathcal{F}_{\beta_t}[\rho_t]$ as a valid Lyapunov function. The dissipation rate is zero if and only if $\nabla f(x) + \frac{1}{\beta_t}\nabla\log\rho_t(x) = 0$ for all $x$ in the support of $\rho_t$.

At equilibrium, when $\frac{\partial\rho_t}{\partial t} = 0$, equation \eqref{eq:continuity} implies that the probability flux vanishes, leading to:
\begin{equation}
	\rho_t(x)\left(-\nabla f(x) - \frac{1}{\beta_t}\nabla\log\rho_t(x)\right) = 0.
\end{equation}
Assuming $\rho_t(x) > 0$ almost everywhere, this reduces to:
\begin{equation}
	-\nabla f(x) - \frac{1}{\beta_t}\nabla\log\rho_t(x) = 0,
	\label{eq:equilibrium_condition}
\end{equation}
which is equivalent to $\nabla f(x) + \frac{1}{\beta_t}\nabla\log\rho_t(x) = 0$, consistent with the condition for zero dissipation in equation \eqref{eq:free_energy_dissipation}.

Rearranging equation \eqref{eq:equilibrium_condition} gives $\nabla\log\rho_t(x) = -\beta_t\nabla f(x)$. Integrating both sides with respect to $x$ yields:
\begin{equation}
	\log\rho_t(x) = -\beta_t f(x) + C,
\end{equation}
where $C$ is an integration constant. Exponentiating produces:
\begin{equation}
	\rho_t(x) = e^C \cdot e^{-\beta_t f(x)}.
\end{equation}

Normalizing to obtain a valid probability density function gives the equilibrium distribution:
\begin{equation}
	\rho_{\text{eq}}(x) = \frac{1}{Z_{\beta_t}} e^{-\beta_t f(x)},
	\label{eq:boltzmann_distribution}
\end{equation}
where $Z_{\beta_t} = \int_{\mathcal{X}} e^{-\beta_t f(x)} dx$ is the partition function ensuring $\int \rho_{\text{eq}}(x) dx = 1$.

This equilibrium distribution is precisely the Boltzmann distribution with energy function $f(x)$ (for minimization). The parameter $\beta_t$ controls the concentration of the distribution: when $\beta_t$ is small, the distribution is diffuse and exploratory; when $\beta_t$ is large, the distribution concentrates on regions where $f(x)$ is minimized.

To establish convergence to this equilibrium, we examine the behavior of the free energy functional at the equilibrium distribution. Substituting equation \eqref{eq:boltzmann_distribution} into equation \eqref{eq:free_energy} gives:
\begin{equation}
	\mathcal{F}_{\beta_t}[\rho_{\text{eq}}] = \frac{1}{Z_{\beta_t}}\int f(x) e^{-\beta_t f(x)} dx - \frac{1}{\beta_t}S[\rho_{\text{eq}}].
\end{equation}

Using the properties of the Boltzmann distribution, we can show that for the specific form $\rho_{\text{eq}}(x) = e^{-\beta_t f(x)}/Z_{\beta_t}$, the entropy is given by:
\begin{equation}
	S[\rho_{\text{eq}}] = \log Z_{\beta_t} + \beta_t \mathbb{E}_{\rho_{\text{eq}}}[f(x)].
	\label{eq:entropy_boltzmann}
\end{equation}

Substituting equation \eqref{eq:entropy_boltzmann} into the free energy expression yields:
\begin{equation}
	\mathcal{F}_{\beta_t}[\rho_{\text{eq}}] = -\frac{1}{\beta_t}\log Z_{\beta_t}.
	\label{eq:free_energy_minimum}
\end{equation}

This represents the minimum possible value of the free energy functional for fixed $\beta_t$. Indeed, for any distribution $\rho$, we have:
\begin{equation}
	\mathcal{F}_{\beta_t}[\rho] - \mathcal{F}_{\beta_t}[\rho_{\text{eq}}] = \frac{1}{\beta_t}D_{\text{KL}}(\rho \|\rho_{\text{eq}}) \geq 0,
	\label{eq:free_energy_gap}
\end{equation}
where $D_{\text{KL}}$ is the Kullback-Leibler divergence, which is non-negative.

The convergence argument proceeds as follows. Equation \eqref{eq:free_energy_dissipation} shows that the free energy decreases monotonically along the trajectory. Combined with the lower bound in equation \eqref{eq:free_energy_gap}, this ensures convergence to the equilibrium distribution. Moreover, as $\beta_t$ increases over time (a process called annealing), the equilibrium distribution becomes increasingly concentrated on the global minimum of $f(x)$.

In the limit $\beta_t \to \infty$, the Boltzmann distribution converges weakly to a Dirac delta distribution centered on the global minimum:
\begin{equation}
	\lim_{\beta_t \to \infty} \rho_{\text{eq}}(x) = \delta(x - x^*),
	\label{eq:dirac_limit}
\end{equation}
where $x^* = \arg\min_{x \in \mathcal{X}} f(x)$, assuming the minimum is unique. For multiple global minima, the distribution converges to a mixture of Dirac deltas at each global minimum.

The mathematical rigor of this convergence is supported by the fact that the dynamics constitute a gradient flow in the space of probability measures equipped with the Wasserstein metric. The free energy functional is geodesically convex under appropriate conditions on $f(x)$, ensuring exponential convergence to equilibrium. The discrete-time implementation with finite particle count introduces approximation errors, but these can be controlled through the choice of kernel bandwidth in density estimation and through the learning rate in the discretization of equation \eqref{eq:particle_update}.

The WE thus provides a principled approach to global optimization that combines gradient-based local search with entropy-driven exploration. The theoretical guarantee of convergence to the Boltzmann distribution ensures that, with appropriate annealing of $\beta_t$, the algorithm will eventually concentrate its search effort on the globally optimal regions of the search space, while maintaining diversity through the entropy term to escape local optima during the initial phases of optimization.

\subsection*{A.4 Performance evaluation metrics}

Two physically grounded metrics are employed to evaluate algorithmic performance: the average differential entropy \(S\) and the free energy \(\mathcal{F}\). The entropy \(S\) quantifies the configurational diversity of the final population. For a final population \(\{\mathbf{x}_i\}_{i=1}^N\) with \(\mathbf{x}_i \in \mathbb{R}^d\), the entropy is estimated via multivariate Gaussian Kernel Density Estimation (KDE):

\begin{equation}
	S = -\frac{1}{N}\sum_{i=1}^N \log \hat{\rho}(\mathbf{x}_i),
	\label{eq:entropy_kde}
\end{equation}

where \(\hat{\rho}(\mathbf{x})\) is the KDE estimate given by:

\begin{equation}
	\hat{\rho}(\mathbf{x}) = \frac{1}{N h^d} \sum_{j=1}^N K\left(\frac{\mathbf{x} - \mathbf{x}_j}{h}\right),
	\label{eq:kde_estimate}
\end{equation}

with \(K(\cdot)\) being the multivariate Gaussian kernel:

\begin{equation}
	K(\mathbf{u}) = \frac{1}{(2\pi)^{d/2}} \exp\left(-\frac{1}{2}\mathbf{u}^\top\mathbf{u}\right).
	\label{eq:gaussian_kernel}
\end{equation}

The free energy \(\mathcal{F}\) is computed based on the thermodynamic definition:
\begin{equation}
\mathcal{F} = U - \frac{1}{\beta} S,
\label{eq:free_energy_def}
\end{equation}
where \(U\) represents the potential energy, which corresponds to the function value  of the final population, and \(S\) is the entropy calculated via Eq.~\eqref{eq:entropy_kde}. The inverse temperature parameter \(\beta\) controls the trade-off between potential minimization and entropy maximization. Lower free energy indicates a more stable, robust system under landscape perturbations, while higher entropy reflects greater population diversity.

As additional evidence for WE maintaining the most diverse population throughout optimization, we compute the population diversity at each iteration using the formula below:
\begin{equation}
\text{Diversity} = \frac{1}{D} \sum_{j=1}^{D} \left( \frac{1}{N} \sum_{i=1}^{N} |x_{ij} - \text{median}_j| \right),
\end{equation}
where \(N\) is the population size, \(D\) is the problem dimension, \(x_{ij}\) is the \(j\)-th dimension of the \(i\)-th individual, and \(\text{median}_j\) is the median value across all individuals in the \(j\)-th dimension. This metric measures the average absolute deviation from the median across all dimensions, providing a direct measure of population spread.

\subsection*{A.5 Sigmoid Normalization for Visual Comparison}

To facilitate intuitive visual comparison of algorithm performance across diverse test functions with varying metric scales, we apply sigmoid normalization to the Free Energy ($\mathcal{F}$) and Entropy ($S$).

For each function-metric combination (table row) with raw values $\{v_j\}_{j=1}^{M}$ from $M$ algorithms, the normalized value $\tilde{v}_j$ is computed as:
\begin{equation}
\tilde{v}_j = \frac{1}{1 + \exp\left(-\frac{v_j - \mu}{\sigma}\right)},
\end{equation}
where $\mu$ and $\sigma$ are the mean and standard deviation of $\{v_j\}$ for that specific row.

This transformation maps values to $(0,1)$ while preserving their relative ranking. It emphasizes performance differences relative to the group distribution, enabling clearer identification of superior algorithms (closer to 0 for $\mathcal{F}$, closer to 1 for $S$) across functions with vastly different raw metric ranges. The normalization enhances visual interpretability without altering statistical conclusions.